\documentclass[10pt,twocolumn,letterpaper]{article}
\usepackage{cvpr}              % To produce the CAMERA-READY version
\usepackage{graphicx}
\usepackage{amsmath}
\usepackage{amssymb}
\usepackage{booktabs}
\usepackage{amsfonts}
\usepackage{multirow}
\usepackage{wrapfig}
\RequirePackage{caption}
\usepackage{float}
\usepackage{color}
\usepackage[table]{xcolor}
\definecolor{green1}{RGB}{220,20,1}
\definecolor{red1}{RGB}{26,153,25}
\usepackage[accsupp]{axessibility}  % Improves PDF readability for those with disabilities.

\usepackage[pagebackref,breaklinks,colorlinks]{hyperref}

\usepackage[capitalize]{cleveref}
\crefname{section}{Sec.}{Secs.}
\Crefname{section}{Section}{Sections}
\Crefname{table}{Table}{Tables}
\crefname{table}{Table}{Tabs.}

\def\cvprPaperID{484} % *** Enter the CVPR Paper ID here
\def\confName{CVPR}
\def\confYear{2023}

\begin{document}

\title{Learning Semantic-Aware Knowledge Guidance for 

Low-Light Image Enhancement}

\author{Yuhui Wu$^{1}$, Chen Pan$^{1}$, Guoqing Wang$^{1}$\thanks{Corresponding author.}, Yang Yang$^{1}$, Jiwei Wei$^{1}$, Chongyi Li$^{2}$, Heng Tao Shen$^{1}$\\
$^{1}$Center for Future Media, University of Electronic Science and Technology of China, China\\
$^{2}$S-Lab, Nanyang Technological University, Singapore\\
{\tt\small wuyuhui132@gmail.com; panchen0103@163.com; gqwang0420@hotmail.com;}\\
{\tt\small dlyyang@gmail.com; mathematic6@gmail.com; chongyi.li@ntu.edu.sg; shenhengtao@hotmail.com}
}
% \author[1]{Yuhui Wu},
% \author[1]{Chen Pan},
% \author[1]{Guoqing Wang},
% \author[1]{Yang Yang},
% \author[1]{Jiwei Wei},
% \author[2]{Chongyi Li},
% \author[1]{Hengtao Shen}

% % University of Electronic Science and Technology of China\\

% \address{University of Electronic Science and Technology of China}
% \address{Nanyang Technological University}
% \author[1]{Yuhui Wu}
% \author[1]{Chen Pan}
% \author[1]{Guoqing Wang \thanks{Corresponding author: email@mail.com}}
% \author[1]{Yang Yang}
% \author[1]{Jiwei Wei}
% \author[2]{Chongyi Li}
% \author[1]{Hengtao Shen}
% \affil[1]{University of Electronic Science and Technology of China}
% \affil[2]{Nanyang Technological University}

% % 使用 \thanks 定义通讯作者

% \renewcommand*{\Affilfont}{\small\it} % 修改机构名称的字体与大小
% \renewcommand\Authands{ and } % 去掉 and 前的逗号
% \date{} % 去掉日期

\twocolumn[{%
\renewcommand\twocolumn[1][]{#1}%
\maketitle
\vspace{-1.3cm} 
\begin{figure}[H]
    \hsize=\textwidth 
    \setlength{\abovecaptionskip}{0.1cm}
    \setlength{\belowcaptionskip}{-0.25cm}
    \centering
    \includegraphics[width=\textwidth]{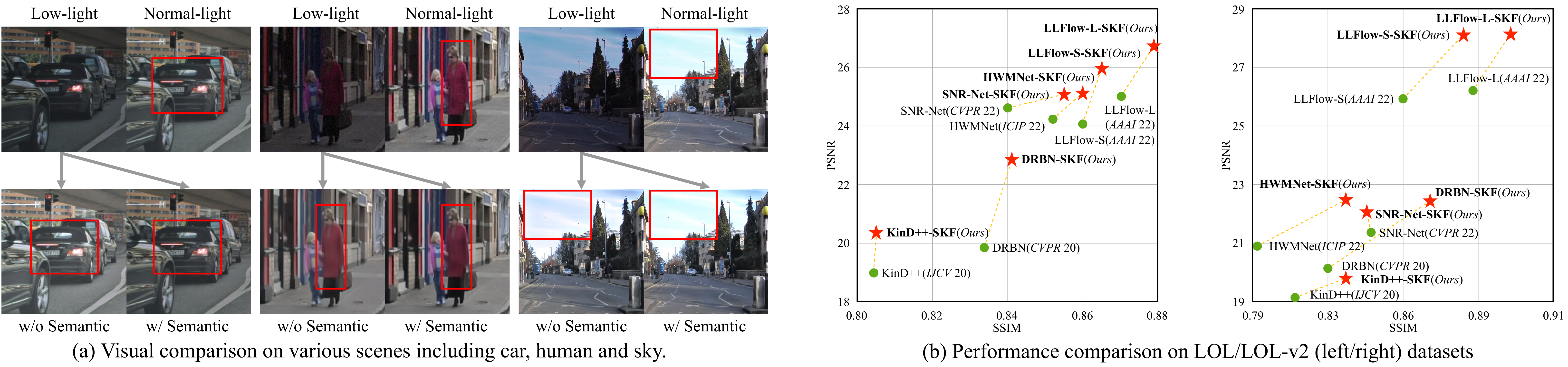}
    \caption{\textbf{Motivation and superiority.} (a) The enhancement results (bottom row) without semantic priors show color deviations (\eg, the black car turns gray). (b) Our SKF provides remarkable performance boost on LOL/LOL-v2 datasets in terms of PSNR/SSIM.}
    \label{fig:motivation}
\end{figure}
}]

% \begin{figure*}[ht]
%   \centering
%   \includegraphics[width=\linewidth]{latex/效果对比+Motivation.pdf}
%   \setlength{\abovecaptionskip}{-0.3cm}
%   \setlength{\belowcaptionskip}{-0.5cm}
%   \caption{Our SKF provides remarkable performance boost for baseline methods on both LOL and LOL-v2 datasets in terms of PSNR/SSIM.}
%   \label{fig:oneco}
% \end{figure*}

%%%%%%%%% ABSTRACT
\begin{abstract}
\vspace{-0.4cm}
Low-light image enhancement (LLIE) investigates how to improve illumination and produce normal-light images. The majority of existing methods improve low-light images via a global and uniform manner, without taking into account the semantic information of different regions. Without semantic priors, a network may easily deviate from a region's original color. To address this issue, we propose a novel semantic-aware knowledge-guided framework (SKF) that can assist a low-light enhancement model in learning rich and diverse priors encapsulated in a semantic segmentation model. We concentrate on incorporating semantic knowledge from three key aspects: a semantic-aware embedding module that wisely integrates semantic priors in feature representation space, a semantic-guided color histogram loss that preserves color consistency of various instances, and a semantic-guided adversarial loss that produces more natural textures by semantic priors. Our SKF is appealing in acting as a general framework in LLIE task. Extensive experiments show that models equipped with the SKF significantly outperform the baselines on multiple datasets and our SKF generalizes to different models and scenes well.
The code is available at \href{https://github.com/langmanbusi/Semantic-Aware-Low-Light-Image-Enhancement}{Semantic-Aware-Low-Light-Image-Enhancement}
% Codes and models are submitted in the \textcolor[rgb]{0.75,0.16,0.26}{supplementary material}.
% Code and model will be released.
%The codes and more results will be publicly available after paper acceptance.

% The code is available at \href{https://github.com/Anonymousstudy1/SKF-LLIE}{https://github.com/Anonymousstudy1/SKF-LLIE}

\end{abstract}

%%%%%%%%% BODY TEXT
\vspace{-0.6cm}
\section{Introduction}
\label{sec:intro}
\vspace{-0.2cm}
In real world, low-light imaging is fairly common due to unavoidable environmental or technical constraints such as insufficient illumination and limited exposure time. 
Low-light images not only have poor visibility for human perception, but also are unsuitable for subsequent multimedia computing and downstream vision tasks designed for high-quality images~\cite{deng2019facede, ren2015objectde, fu2019semantic}. Thus, low-light image enhancement (LLIE) is proposed to reveal buried details in low-light images and avoid degraded performance in subsequent vision tasks. Mainstream traditional methods for LLIE include Histogram Equalization-based methods~\cite{abdullah2007histogramequal} and Retinex model-based methods~\cite{jobson1997retinex}. 

Recently, many  deep learning-based LLIE methods have proposed, such as end-to-end frameworks~\cite{moran2020deeplpf,yang2020drbn,xu2020learning,fan2022hwmnet,xu2022snr,dong2022abandoning} and Retinex-based frameworks~\cite{Chen2018Retinex,zhang2019kind,yang2021sparse,zhang2021kindplus,liu2021ruas,wu2022uretinexnet,wang2022llflow}. 
Benefiting from their ability in modeling the mapping between the low-light and high-quality image, deep LLIE methods commonly achieve better results than traditional approaches. However, existing methods typically improve low-light images globally and uniformly, without taking into account the semantic information of different regions, which is crucial for enhancement. As shown in~\cref{fig:motivation}\textcolor{red}{(a)}, a network that lacks the utilization of semantic priors can easily deviate from a region's original hue~\cite{li2021lliesurvey}. Furthermore, studies have demonstrated the significance of incorporating semantic priors into low-light enhancement. Fan \etal.~\cite{fan2020integrating} utilize semantic map as prior and incorporated it into the feature representation space, thereby enhancing image quality. Rather than relying on optimizing intermediate features, Zheng \etal.~\cite{zheng2022semanticzeroLLIE} adopt a novel loss to guarantee the semantic consistency of the enhanced images. These methods successfully combine the semantic priors with LLIE task, demonstrating the superiority of semantic constraints and guidance. However, their methods fail to fully exploit the knowledge that semantic segmentation networks can provide, limiting the performance gain by semantic priors. Furthermore, the interaction between segmentation and enhancement is designed for specific methods, limiting the possibility of incorporating semantic guidance into LLIE task. Hence, we wonder two questions: \textit{1. How can we obtain various and available semantic knowledge? 2. How does semantic knowledge contribute to image quality improvement in LLIE task?} 

We attempt to answer the first question.
First, a semantic segmentation network pre-trained on large-scale datasets is introduced as a semantic knowledge bank (SKB). The SKB can provide richer and more diverse semantic priors to improve the capability of enhancement networks. 
Second, according to previous works~\cite{fan2020integrating,zheng2022semanticzeroLLIE,jung2021semanticdepth1}, the available priors provided by the SKB primarily consist of intermediate features and semantic maps. Once training a LLIE model, the SKB yields above semantic priors and guides the enhancement process. The priors can not only refine image features by employing techniques like affinity matrices, spatial feature transformations~\cite{wang2018sft}, and attention mechanisms, but also guide the design of objective functions by explicitly incorporating regional information into LLIE task~\cite{liang2022SCLLLE}.

Then we try to answer the second question.
We design a series of novel methods to integrate semantic knowledge into LLIE task based on the above answers, formulating in a novel semantic-aware knowledge-guided framework (SKF). First, we use the High-Resolution Network~\cite{wang2020hrnet} (HRNet) pre-trained on the PASCAL-Context dataset~\cite{mottaghi2014pascalcontext} as the previously mentioned SKB. In order to make use of intermediate features, we develop a semantic-aware embedding (SE) module. It computes the similarity between the reference and target features and employs cross-modal interactions between heterogeneous representations. As a result, we quantify the semantic awareness of image features as a form of attention and embed semantic consistency in enhancement network.

Second, some methods~\cite{zhang2022colordccnet,kim2022colorhistloss} propose to optimize image enhancement using color histogram in order to preserve the color consistency of the image rather than simply enhancing the brightness globally. The color histogram, on the other hand, is still a global statistical feature that cannot guarantee local consistency. Hence, we propose a semantic-guided color histogram (SCH) loss to refine color consistency. Here, we intend to make use of local geometric information derived from the scene semantics and global color information derived from the content. In addition to guarantee original color of the enhanced image, it can also add spatial information to the color histogram, performing a more nuanced color recovery.

Third, existing loss functions are not well aligned with human perception and fail to capture an image's intrinsic signal structure, resulting in unpleasing visual results. To improve visual quality, EnlightenGAN~\cite{jiang2021enlightengan} employs global and local image-content consistency and randomly chooses the local patch. However, the discriminator do not know where the regions are likely to be `fake'. Thus, we propose a semantic-guided adversarial (SA) loss. Specifically, the ability of the discriminator is improved by using segmentation map to determine the fake areas, which can improve the image quality further.
% \lichongyi{we may want to mention figure 1(b) in somewhere of Introduction.}

The main contributions of our work are as follows:
\vspace{-0.2cm}
\begin{itemize}
\setlength{\itemsep}{0pt}
\setlength{\parskip}{2pt}
% \setlength{\leftmargin}{0pt}
    %\item We propose a semantic-aware knowledge-guided framework (SKF) to help existing methods exceed their limitation by jointly maintaining color consistency and improving image quality.
    \item We propose a semantic-aware knowledge-guided framework (SKF) to boost the performance of existing methods by jointly maintaining color consistency and improving image quality.
    \item We propose three key techniques to take full advantage of semantic priors provided by semantic knowledge bank (SKB): semantic-aware embedding (SE) module, semantic-guided color histogram (SCH) loss, and semantic-guided adversarial (SA) loss.
    \item We conduct experiments on LOL/LOL-v2 datasets and unpaired datasets. The experimental results demonstrate large performance improvements by our SKF, verifying its effectiveness in resolving the LLIE task.
\end{itemize}

\begin{figure*}[ht]
  \centering
   \includegraphics[width=\linewidth]{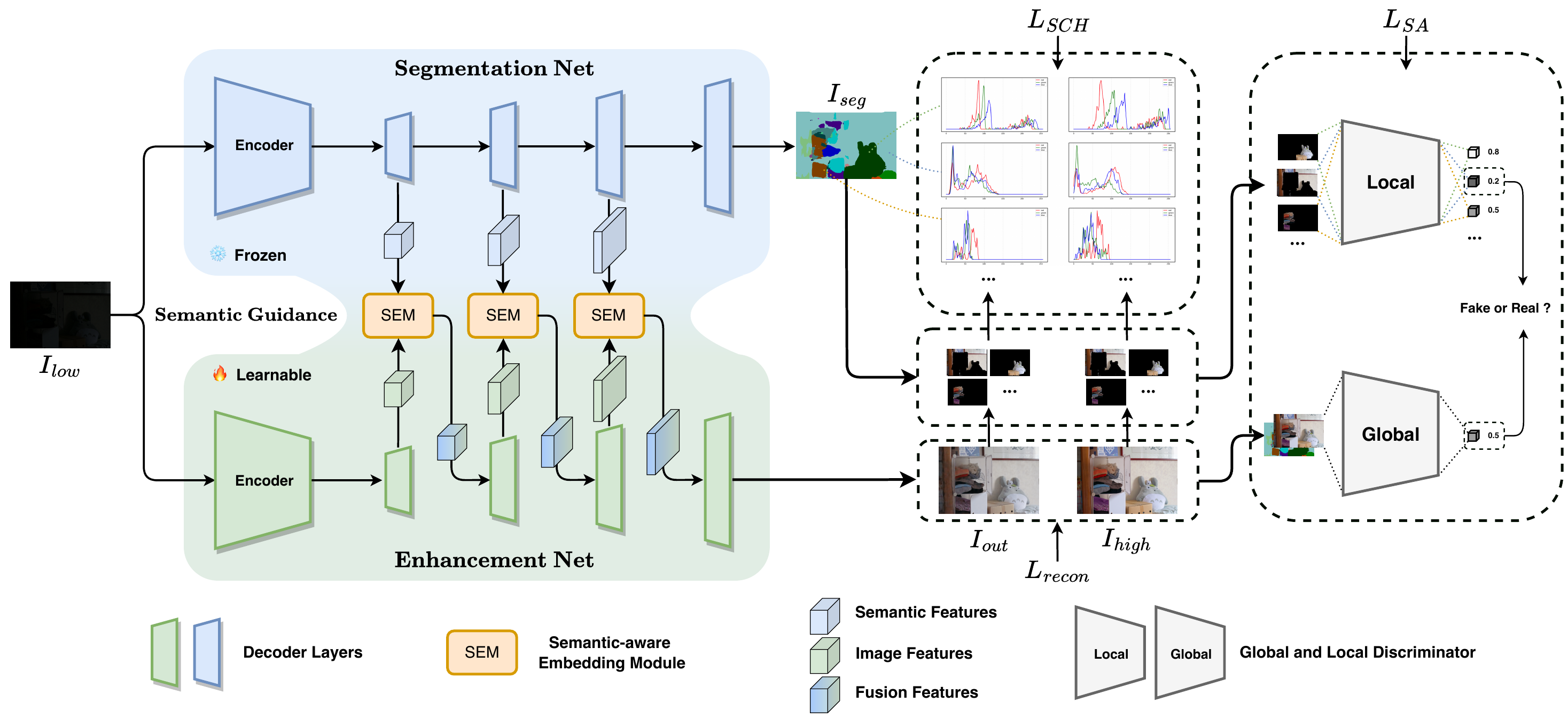}
   \setlength{\abovecaptionskip}{-0.3cm}
   \setlength{\belowcaptionskip}{-0.6cm}
   \caption{Overview of our Semantic-aware Knowledge-guided Framework (SKF). With a pre-trained Segmentation Net, our SKF utilizes semantic priors to improve the enhancement process in two aspects: \textbf{(a)} In feature-level, the multi-scale semantic-aware embedding modules enable cross-modal interactions between semantic features and image features in representation space. \textbf{(b)} In loss-level, the semantic segmentation result is introduced into the computation of color histogram loss and adversarial loss as a guidance.}
   \label{fig:framwork}
\end{figure*}

\vspace{-0.5cm}
%-------------------------------------------------------------------------
\section{Related Work}
\label{sec:related}
% \lichongyi{The related work is a little overlength. We can make it shorter and the key point is to explain the differences between our method and previous related works in the related work. }
\vspace{-0.1cm}
\subsection{Low-light Image Enhancement}
\vspace{-0.05cm}
\textbf{Traditional methods.} Traditional methods for low-light enhancement include Histogram Equalization-based methods~\cite{abdullah2007histogramequal} and Retinex model-based methods~\cite{jobson1997retinex}. The former improve low-light images by extending the dynamic range. The latter decompose a low-light image into reflection and illumination maps and the reflection component is treated as the enhanced image. Such model-based methods require explicit priors to fit data well, but designing proper priors for various scenes is difficult~\cite{wu2022uretinexnet}.

\textbf{Learning-based methods.} Recent deep learning-based methods show promising results~\cite{Chen2018Retinex, zhao2021retinexdip, zhang2019kind, zhang2021kindplus, wu2022uretinexnet, liu2021ruas, jiang2022drgn}. We can further divide existing designs into Retinex-based methods and end-to-end methods. Retinex-based methods use deep network to decompose and enhance an image. Wei \etal. proposed a two-stage Retinex-based method called Retinex-Net~\cite{Chen2018Retinex}. Inspired by Retinex-Net, Zhang \etal. proposed two refined methods, called KinD~\cite{zhang2019kind} and KinD++~\cite{zhang2021kindplus}. Recently, Wu \etal.~\cite{wu2022uretinexnet} proposed a novel deep unfolding Retinex-based network to further integrate the strengths of model-based and learning-based methods.

In comparison to Retinex-based method, end-to-end methods directly learning an enhanced result~\cite{zhu2020eemefn, xu2020learning,lim2020dslr, moran2020deeplpf, zheng2021utvnet, dong2022abandoning, wang2022llflow, fan2022hwmnet, ma2022sci,dudhane2022burst, zamir2022mirnetv2, tu2022maxim, xu2022snr}. Lore \etal.~\cite{lore2017llnet} made the first attempt by proposing a deep autoencoder named Low-Light Net (LLNet). Later on, various end-to-end methods are proposed. Physics-based concepts, \eg Laplacian pyramid~\cite{lim2020dslr}, local parametric filter~\cite{moran2020deeplpf}, Lagrange multiplier~\cite{zheng2021utvnet}, De-Bayer-Filter~\cite{dong2022abandoning}, normalization flow~\cite{wang2022llflow} and wavelet transform~\cite{fan2022hwmnet}, are proposed to improve model interpretability and lead to visually pleasing results. In~\cite{yang2020drbn, jiang2021enlightengan, jin2022nightimage}, adversarial learning is introduced to capture the visual properties. In~\cite{guo2020zerodce}, the light enhancement is creatively formulated as a task of image-specific curve estimation using zero-shot learning. In~\cite{yang2022adaint, kim2022colorhistloss, zhang2022colordccnet}, 3D lookup table and color histogram are utilized to preserve the color consistency. However, existing designs focus on optimizing enhancement process, while ignoring the semantic information of different regions. In contrast, we design a SKF with three key techniques to explore the potential of semantic priors and thus produce visually pleasing enhanced results.

\vspace{-0.1cm}
\subsection{Semantic-Guided Methods}
\vspace{-0.1cm}
Recently, semantic-guided methods prove the reliability of semantic priors. These methods could be divided into two kinds: loss-level semantic-guided methods and feature-level semantic-guided methods. 

\textbf{Loss-level semantic-guided methods.} In order to make use of semantic priors, some works focus on utilizing semantic-aware losses as extra objective functions of the original vision tasks. In image denoising~\cite{liu2018semanticdenoise}, image super-resolution~\cite{aakerberg2022semanticsr}, low-light image enhancement~\cite{zheng2022semanticzeroLLIE}, researchers directly utilized semantic segmentation loss as an extra constrain to guide the training process. Furthermore, Liang \etal.~\cite{liang2022SCLLLE} better maintained the details of the images by using a semantic brightness consistency loss.

\textbf{Feature-level semantic-guided methods.} In comparison to loss-level semantic-guided methods, feature-level semantic-guided methods concentrate on extracting intermediate features from semantic segmentation network and introduce semantic priors in feature representation space to combine with image features. Similar works have been done on image restoration~\cite{li2020dfdnet}, image deraining~\cite{li2022close}, image super-resolution~\cite{wang2018sft}, low-light image enhancement~\cite{fan2020integrating}, depth estimation~\cite{guizilini2019semanticdepth3, jung2021semanticdepth1}.

Existing semantic-guided methods are limited by the insufficient interaction between semantic priors and original tasks. Hence, we propose a semantic-aware framework to fully exploit semantic information both on loss-level and feature-level, including two semantic-guided losses and a semantic-aware embedding module. Specifically, comparing to semantic-guided methods in LLIE task~\cite{fan2020integrating,zheng2022semanticzeroLLIE,liang2022SCLLLE}, our SKF is appealing in acting as a general framework. 

% \lichongyi{we need a paragraph to explain the differences or advantages of our method.}
\vspace{-0.1cm}
\section{Method}
\vspace{-0.1cm}
% In this section, we introduce the motivation and details of SKF (see \cref{fig:framwork}), which has three key components (\textit{i.e.}, SE module, SCH loss and SA loss).
% \lichongyi{can be removed}

\subsection{Motivation and Overview}
\vspace{-0.1cm}
Illumination enhancement is the process of making an underexposed image look better by adjusting the lighting, eliminating noise, and restoring lost details. Semantic priors can provide a wealth of information for improving the enhancement performance. Specifically, semantic priors can help reformulate the existing LLIE methods as a region-aware enhancement framework. In particular, the novel model will blur noises on smooth regions in a simple way, such as skies, whereas being careful on regions with rich details, such as indoor scene. Furthermore, combining with semantic prior, the color consistency of enhanced image will be carefully preserved. A network that lacks access to semantic priors can easily deviate from a region's original hue~\cite{li2021lliesurvey}. Existing low-light enhancement methods, however, ignore the importance of semantic information and thus have limited capability.

In this paper, we propose a novel SKF, jointly optimizing image features, maintaining regional color consistency and improving image quality. As shown in~\cref{fig:framwork}, semantic priors are provided by SKB and integrated into LLIE task by three key components: SE module, SCH loss and SA loss. 

\textbf{Problem definition of semantic-aware LLIE.} Given a low-light image $I_l\in\mathbb{R}^{W\times H\times 3}$ with width $W$ and height $H$. Combining with semantic segmentation, the LLIE process can be modeled as two functions, first: 
\vspace{-0.09cm}
\begin{equation}
  M = \mathbf{F}_{segment}(I_{l}; \theta_s),
  \label{eq:def1}
  \vspace{-0.15cm}
\end{equation}

where $M$ is the semantic prior, including segmentation result and intermediate features with multi-scale dimensions. $\mathbf{F}_{segment}$ represents the pre-trained semantic segmentation network, acting as the SKB, and $\theta_s$ is frozen in training stage. Then $M$ is used as input: 
\vspace{-0.09cm}
\begin{equation}
  \widehat{I_{h}} = \mathbf{F}_{enhance}(I_{l}, M; \theta_e),
  \label{eq:def2}
  \vspace{-0.09cm}
\end{equation}

where $\widehat{I_{h}}\in\mathbb{R}^{W\times H\times 3}$ is the enhanced result and $\mathbf{F}_{enhance}$ represents the enhancement network. During training stage, $\theta_e$ will be updated by minimizing the objective function with the guidance of $M$ while $\theta_s$ is fixed:
\vspace{-0.09cm}
\begin{equation}
  \widehat{\theta_e} = argmin\mathcal{L}(\widehat{I_{h}}, I_{h}, M),
  \label{eq:def3}
  \vspace{-0.09cm}
\end{equation}

where $I_{h}\in\mathbb{R}^{W\times H\times 3}$ is the ground truth, $\mathcal{L}(\widehat{I_{h}}, I_{h}, M)$ is the objective function of semantic-aware LLIE.

\vspace{-0.1cm}
\subsection{Semantic-Aware Embedding Module}
\begin{figure}[t]
  \centering
   \includegraphics[width=\linewidth]{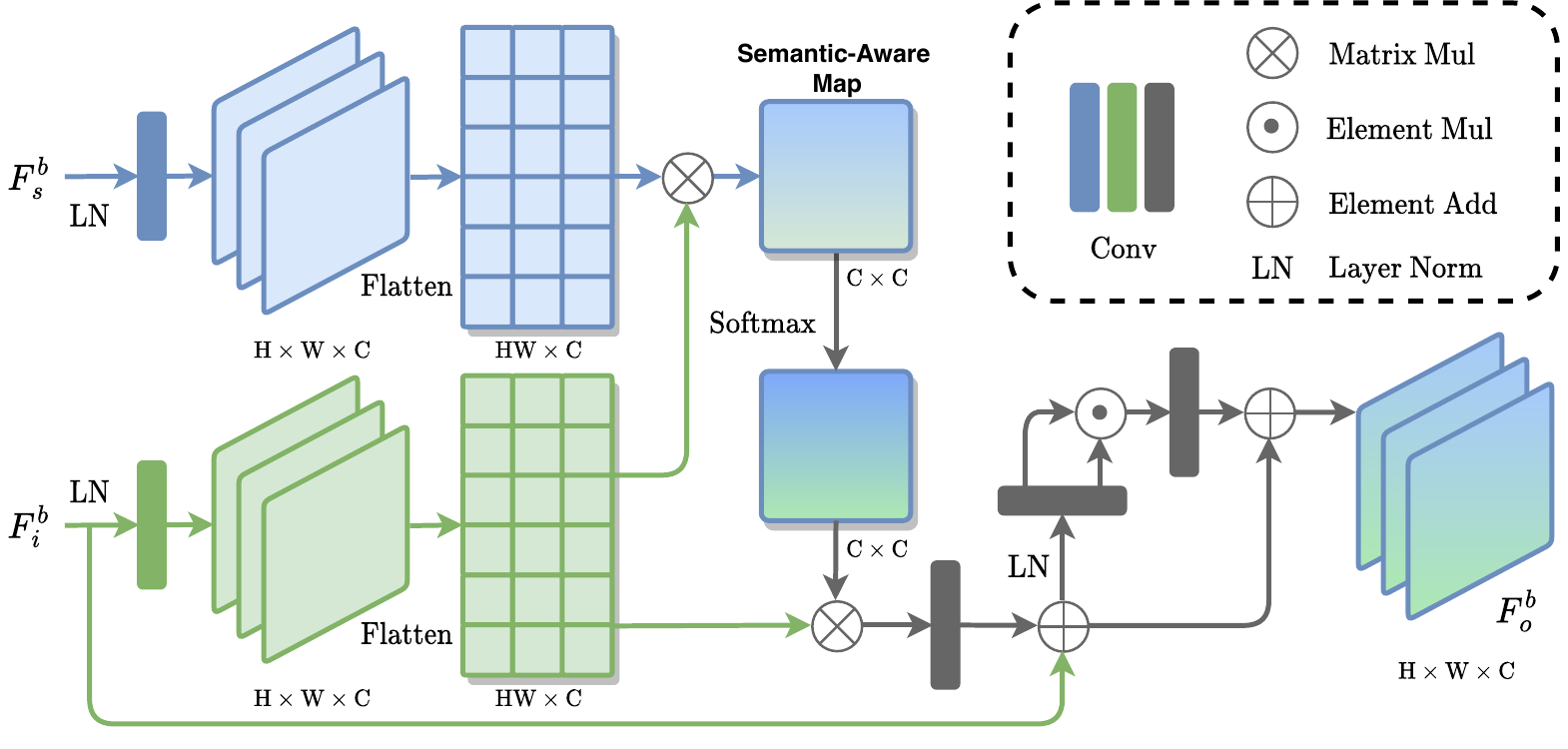}
   \setlength{\abovecaptionskip}{-0.3cm}
   \setlength{\belowcaptionskip}{-0.5cm}
   \caption{Architecture of the semantic-aware embedding (SE) module. At the $ \textit b^{th} $ decoder layer, SE module transforms the image feature map $\textit F_{i}^{b} $ with the semantic feature map $\textit F_{s}^{b} $ and produces the refined output feature $\textit F_{o}^{b} $.}
   \label{fig:msam}
\end{figure}

\vspace{-0.05cm}
When refining image features with the help of semantic priors, another challenge should be particularly considered is the discrepancy between the two sources. To alleviate this issue, we propose the SE module to refine the image feature maps, as illustrated in \cref{fig:msam}. The SE modules are like bridges between Segmentation Net and Enhancement Net (see ~\cref{fig:framwork}), establishing connections between two heterogeneous tasks.

In our framework, we choose HRNet~\cite{wang2020hrnet} as the SKB due to its exceptional performance and make some task-specific modifications. Besides the semantic map, we utilize output features before the representation head as multi-scale semantic priors. For further illustration, three SE modules are shown in~\cref{fig:framwork}, thus we take three semantic/image features ($\textit{F}_s^b / \textit{F}_i^b,b=0,1,2$) with three spatial resolutions ($\textit{H}/2^{4-b}, \textit{W}/2^{4-b}$), where $\textit{H}$ and $\textit{W}$ are the height and width of the input image. The SE module performs a pixel-wise interaction between $\textit{F}_s^b$ and $\textit{F}_i^b$, and gives the final refined feature map $\textit{F}_o^b$. Details of the learning process are provided below.

The SE module computes the semantic awareness of the image features through cross-modal similarity and produces a semantic-aware map. We first apply convolution layers to transform $\textit{F}_s^b$ and $\textit{F}_i^b$ to the same dimension. Next, inspired by Restormer~\cite{zamir2022restormer}, we adopt a transposed-attention mechanism to compute the attention map with a low computational cost. Hence, the semantic-aware attention map is described as follows:
\vspace{-0.1cm}
\begin{equation}
  A^b = Softmax\left(W_k(\textit{F}_i^{b}) \times W_q(\textit{F}_s^{ b})/\sqrt{C}\right),
  \label{eq:msa1}
  \vspace{-0.1cm}
\end{equation}

where $W_k(\cdot)$ and $W_q(\cdot)$ are convolution layers, $LN$ is layer normalization, $C$ is channel of features. Here, $A^b\in\mathbb{R}^{C\times C}$ indicates the semantic-aware attention map, which represents the interrelationship between $\textit{F}_i^b$ and $\textit{F}_s^b$. Then we use $A^b$ to fabricate image feature $\textit{F}_i^b$ as follows:
\vspace{-0.1cm}
\begin{equation}
  \textit{F}_o^b = FN(W_v(\textit{F}_i^b) \times A^b + \textit{F}_i^b),
  \label{eq:msa2}
  \vspace{-0.1cm}
\end{equation}
where $FN$ denotes feed-forward network, $\textit{F}_o^b$ is the final refined feature map of $b^{th}$ SE module and becomes the input of ${(b\mathit{+}1)}^{th}$ layer of the Enhancement Net decoder.

\subsection{Semantic-Guided Color Histogram Loss}
\label{subsec:scloss}

Color histogram carries crucial underlying image statistics and is profitable for learning color representations. DCC-Net~\cite{zhang2022colordccnet} uses PCE modules with affinity matrix to match the color histogram and content in feature-level, therefore retaining color consistency of enhanced image. However, color histogram describes a global statistic information, differences in color characteristics between various instances are eliminated. Thus, we propose an intuitive way to achieve local color adjustment, \textit{i.e.}, semantic-guided color histogram (SCH) loss, as shown in~\cref{fig:framwork}. It focuses on adjusting color histogram of each instance, thereby retaining more detailed color information. 

The semantic map is firstly used to divide enhanced result into image patches with different instance labels. Each patch includes a single instance with the same label. Hence, the process of producing patches are defined as follows:
\vspace{-0.1cm}
\begin{equation}
  P = \{P^{0}, P^{1}, \dotsc, P^{class}\}, \quad P^{c} = I_{out} \odot I_{seg}^{c},
  \label{eq:scloss1}
  \vspace{-0.1cm}
\end{equation}
% \vspace{-0.4cm}
% \begin{equation}
%   {\rm where} \quad P^{c} = I_{out} \odot I_{seg}^{c},
%   \label{eq:scloss2}
% \end{equation}
where $\odot$ is the dot product, $I_{out}$ denotes the enhanced result, $I_{seg}^{c}$ denotes the $c^{th}$ channel of the one-hot semantic map, $P^c\in\mathbb{R}^{W\times H\times 3}$ denotes the $c^{th}$ image patch, $P$ denotes the group of all the patches. 

Due to the discrete nature of color histogram, we approximate the differentiable version inspired by Kernel Density Estimation~\cite{avi2020deephist} for model training. Considering the prediction error of the semantic result, pixels that are close to the boundary are not considered. We refine the patch group $P$ to $P'$ without edge pixels, mitigating effects of misclassification. In the case of R channel of the $c^{th}$ image patch ${P^c}'(R)$, the estimation process is defined as follows:
\vspace{-0.1cm}
\begin{equation}
  x_{ij}^h=x_j-\frac{i-0.5}{255}{\rm ,\quad} x_{ij}^l=x_j-\frac{i+0.5}{255},
  \label{eq:scloss3}
  \vspace{-0.1cm}
\end{equation}
where $x_j$ denotes $j^{th}$ pixel in ${P^c}'(R)$, $i\in [0,255]$ denotes pixel intensity. $x_{ij}^h$ and $x_{ij}^l$ represent higher anchor and lower anchor respectively, which are key variables to estimate histogram as follows:
\vspace{-0.15cm}
\begin{flalign}
& H_i^c \mathit{=} \sum\limits_{j} \left(Sigmoid(\alpha \cdot x_{ij}^h) \mathit{-} Sigmoid(\alpha \cdot x_{ij}^l)\right), &
\label{eq:scloss4}
\end{flalign}
\vspace{-0.6cm}
\begin{equation}
  H^c = {\{i, H_i^c\}}_{i=0}^{255},
  \label{eq:scloss4_1}
\end{equation}
where $H^c$ denotes the differentiable histogram of ${P^c}'(R)$, $H_i^c$ denotes estimated number of pixels with intensity value $i$. $\alpha$ is a scaling factor, we experimentally set it to 400 for better estimation. The difference between results of two $Sigmoid(\cdot)$ denotes the contribution of $x_j$ to the number of pixels with intensity value $i$. Specifically, when $x_j$ exactly equal to $i$, the difference is 1, \textit{i.e.}, $x_j$ adds 1 to $H_i^c$.

Finally, we apply $l_1$ loss to constrain the estimated differentiable histogram. Therefore, the SCH loss can be described as follows:
\vspace{-0.1cm}
\begin{equation}
  \mathcal{L}_{SCH} = \sum\limits_{c} \parallel H^c(\hat{I_h}) - H^c(I_h) \parallel_1,
  \label{eq:scloss5}
  \vspace{-0.2cm}
\end{equation}
where $\hat{I_h}$ and $I_h$ denote output and groundtruth respectively, $H^c(\cdot)$ denotes the process of histogram estimation.

\vspace{-0.1cm}
\subsection{Semantic-Guided Adversarial Loss}
\vspace{-0.1cm}
\label{subsec:saloss}

\begin{table*}[ht]
  \centering
  \caption{Quantitative comparison on the LOL~\cite{Chen2018Retinex} and LOL-v2~\cite{yang2021sparse} datasets. $\uparrow$ ($\downarrow$) denotes that, larger (smaller) values lead to better quality. \small \textcolor{green1}{+} \small (\textcolor{red1}{-}) denotes the \textcolor{green1}{improvement} (\textcolor{red1}{reduction}) of performance, corresponding to $\uparrow$ ($\downarrow$). The bold denotes the best.}
  \scalebox{.69}{
    \begin{tabular}{c|c|c|c|c|c|c|c|c|c}
    \toprule
    \multirow{2}[2]{*}{\textbf{Method}} & \multicolumn{4}{c|}{\textbf{LOL}} & \multicolumn{4}{c|}{\textbf{LOL-v2}} & \multirow{2}[2]*{\textbf{Param(M)}} \\
\cmidrule{2-9}          & \textbf{PSNR} $\uparrow$ & \textbf{SSIM} $\uparrow$ & \textbf{LPIPS} $\downarrow$ & \textbf{NIQE} $\downarrow$ & \textbf{PSNR} $\uparrow$ & \textbf{SSIM} $\uparrow$ & \textbf{LPIPS} $\downarrow$ & \textbf{NIQE} $\downarrow$ &  \\
    \midrule
    LIME~\cite{guo2016lime} \scriptsize TIP'16 & 16.760  & 0.560  & 0.350  & -     & 15.240  & 0.470  & -     & -     & - \\
    Zero-DCE~\cite{guo2020zerodce} \scriptsize CVPR'20 & 14.861  & 0.562  & 0.335  & 7.767     & 18.059  & 0.580  & 0.313     & 8.058     & 0.33  \\
    % MIRNet~\cite{zamir2020mirnet} \scriptsize ECCV'20 & 24.140  & 0.830  & 0.130  & -     & 20.020  & 0.820  & -     & -     & 5.90  \\
    EnlightGAN~\cite{jiang2021enlightengan} \scriptsize TIP'21 & 17.483  & 0.652  & 0.322  & 4.684     & 18.640  & 0.677  & 0.309     & 5.089     & 8.64  \\
    ISSR~\cite{fan2020integrating} \scriptsize MM'20 & 18.846  & 0.788  & 0.243  & 5.249  & 16.994  & 0.798  & 0.206     & 5.179     & 12.12  \\
    MIRNet~\cite{zamir2022mirnetv2} \scriptsize PAMI'22 & 24.140  & 0.842  & 0.131  & 4.203     & 20.357  & 0.782  & 0.317     & 5.094     & 5.90  \\
    \midrule
    \midrule
    RetinexNet~\cite{Chen2018Retinex} \scriptsize BMVC'18 & 16.770  & 0.462  & 0.474  & 8.873  & 18.371  & 0.723  & 0.365  & 5.849  & 0.62  \\ % \rowcolor{gray!11}
    \textbf{RetinexNet-SKF(Ours)} & 20.418\,\footnotesize \textcolor{green1}{(+3.648)} & 0.711\,\footnotesize \textcolor{green1}{(+0.249)} & 0.216\,\footnotesize \textcolor{green1}{(+0.258)} & 4.211\,\footnotesize \textcolor{green1}{(+4.662)} & 19.849\,\footnotesize \textcolor{green1}{(+1.478)} & 0.719\,\footnotesize \textcolor{red1}{(-0.004)} & 0.255\,\footnotesize \textcolor{green1}{(+0.110)} & 4.233\,\footnotesize \textcolor{green1}{(+1.616)} & 0.66  \\
    \midrule
    \midrule
    KinD~\cite{zhang2019kind} \scriptsize MM'19 & 20.870  & 0.799  & 0.207  & 5.189  & 17.544  & 0.669  & 0.375  & 6.849  & 8.03  \\ % \rowcolor{gray!11}
    \textbf{KinD-SKF(Ours)} & 21.913\,\footnotesize \textcolor{green1}{(+1.043)} & 0.835\,\footnotesize \textcolor{green1}{(+0.036)} & 0.143\,\footnotesize \textcolor{green1}{(+0.064)} & 5.031\,\footnotesize \textcolor{green1}{(+0.158)} & 19.821\,\footnotesize \textcolor{green1}{(+2.277)} & 0.833\,\footnotesize \textcolor{green1}{(+0.164)} & 0.201\,\footnotesize \textcolor{green1}{(+0.174)} & 4.778\,\footnotesize \textcolor{green1}{(+2.071)} & 8.50  \\
    \midrule
    \midrule
    DRBN~\cite{yang2020drbn} \scriptsize CVPR'20 & 19.860  & 0.834  & 0.155  & 4.793  & 20.130  & 0.830  & 0.147  & 4.961  & 2.21  \\ % \rowcolor{gray!11}
    \textbf{DRBN-SKF(Ours)} & 22.837\,\footnotesize \textcolor{green1}{(+2.977)} & 0.841\,\footnotesize \textcolor{green1}{(+0.007)} & 0.138\,\footnotesize \textcolor{green1}{(+0.017)} & 4.464\,\footnotesize \textcolor{green1}{(+0.329)} & 22.441\,\footnotesize \textcolor{green1}{(+2.311)} & 0.871\,\footnotesize \textcolor{green1}{(+0.041)} & 0.132\,\footnotesize \textcolor{green1}{(+0.015)} & 4.460\,\footnotesize \textcolor{green1}{(+0.501)} & 2.43  \\
    \midrule
    \midrule
    KinD++~\cite{zhang2021kindplus} \scriptsize IJCV'20 & 18.970  & 0.804  & 0.175  & 4.760  & 19.087  & 0.817  & 0.180  & 5.086  & 9.63  \\ % \rowcolor{gray!11}
    \textbf{KinD++-SKF(Ours)} & 20.363\,\footnotesize \textcolor{green1}{(+1.393)} & 0.805\,\footnotesize \textcolor{green1}{(+0.001)} & 0.201\,\footnotesize \textcolor{red1}{(-0.026)} & \textbf{4.142}\,\footnotesize \textcolor{green1}{\textbf{(+0.618)}} & 19.779\,\footnotesize \textcolor{green1}{(+0.692)} & 0.837\,\footnotesize \textcolor{green1}{(+0.020)} & 0.178\,\footnotesize \textcolor{green1}{(+0.002)} & 4.179\,\footnotesize \textcolor{green1}{(+0.907)} & 10.21  \\
    \midrule
    \midrule
    HWMNet~\cite{fan2022hwmnet} \scriptsize ICIP'22 & 24.240  & 0.852  & 0.114  & 5.141  & 20.928  & 0.798  & 0.359  & 5.970  & 66.56  \\ % \rowcolor{gray!11}
    \textbf{HWMNet-SKF(Ours)} & 25.086\,\footnotesize \textcolor{green1}{(+0.846)} & 0.860\,\footnotesize \textcolor{green1}{(+0.008)} & 0.108\,\footnotesize \textcolor{green1}{(+0.006)} & 4.346\,\footnotesize \textcolor{green1}{(+0.795)} & 22.490\,\footnotesize \textcolor{green1}{(+1.562)} & 0.836\,\footnotesize \textcolor{green1}{(+0.038)} & 0.175\,\footnotesize \textcolor{green1}{(+0.184)} & 4.683\,\footnotesize \textcolor{green1}{(+1.288)} & 69.98  \\
    \midrule
    \midrule
    SNR-LLIE-Net~\cite{xu2022snr} \scriptsize CVPR'22 & 24.608  & 0.840  & 0.151  & 5.179   & 21.479  & 0.848  & 0.157     & 4.623     & 39.13 \\ % \rowcolor{gray!11}
    \textbf{SNR-LLIE-Net-SKF(Ours)} & 25.031\,\footnotesize \textcolor{green1}{(+0.552)} & 0.855\,\footnotesize \textcolor{green1}{(+0.015)} & 0.113\,\footnotesize \textcolor{green1}{(+0.038)} & 4.722\,\footnotesize \textcolor{green1}{(+0.457)} & 21.927\,\footnotesize \textcolor{green1}{(+0.448)} & 0.842\,\footnotesize \textcolor{red1}{(-0.006)} & 0.160\,\footnotesize \textcolor{red1}{(-0.003)} & \textbf{3.963}\,\footnotesize \textcolor{green1}{\textbf{(+0.660)}} & 39.44  \\
    \midrule
    \midrule
    LLFlow-S~\cite{wang2022llflow} \scriptsize AAAI'22 & 24.060  & 0.860  & 0.136  & 5.412  & 25.922  & 0.860  & 0.173  & 6.150  & 4.97  \\ % \rowcolor{gray!11}
    \textbf{LLFlow-S-SKF(Ours)} & 25.942\,\footnotesize \textcolor{green1}{(+1.882)} & 0.865\,\footnotesize \textcolor{green1}{(+0.005)} & 0.125\,\footnotesize \textcolor{green1}{(+0.011)} & 5.606\,\footnotesize \textcolor{red1}{(-0.194)} & 28.107\,\footnotesize \textcolor{green1}{(+2.185)} & 0.884\,\footnotesize \textcolor{green1}{(+0.024)} & 0.133\,\footnotesize \textcolor{green1}{(+0.040)} & 5.415\,\footnotesize \textcolor{green1}{(+0.735)} & 5.26  \\
    \midrule
    \midrule
    LLFlow-L~\cite{wang2022llflow} \scriptsize AAAI'22 & 24.999  & 0.870  & 0.117  & 5.582  & 26.200  & 0.888  & 0.137  & 5.406  & 37.68  \\ % \rowcolor{gray!11}
    \textbf{LLFlow-L-SKF(Ours)} & \textbf{26.798}\,\footnotesize \textcolor{green1}{\textbf{(+1.799)}} & \textbf{0.879}\,\footnotesize \textcolor{green1}{\textbf{(+0.009)}} & \textbf{0.105}\,\footnotesize \textcolor{green1}{\textbf{(+0.012)}} & 5.589\,\footnotesize \textcolor{red1}{(-0.007)} & \textbf{28.451}\,\footnotesize \textcolor{green1}{\textbf{(+2.251)}} & \textbf{0.905}\,\footnotesize \textcolor{green1}{\textbf{(+0.017)}} & \textbf{0.112}\,\footnotesize \textcolor{green1}{\textbf{(+0.025)}} & 5.725\,\footnotesize \textcolor{red1}{(-0.319)} & 39.91  \\
    \bottomrule
    \end{tabular}}
  \label{tab:lollolv2}%
  \vspace{-0.3cm}
\end{table*}%

Global and local discriminator is used to encourage more realistic results in image inpainting tasks~\cite{li2017generative,iizuka2017globally}. EnlightenGAN~\cite{jiang2021enlightengan} employs this idea as well, but the local patches are selected randomly instead of focusing on fake regions. Therefore we introduce semantic information to guide the discriminator to focus on intriguing regions. To achieve this, we further refine the global and local adversarial loss function respectively by the segmentation map $I_{seg}$ and image patches $P'$ mentioned in~\cref{subsec:scloss}. Finally, we propose the semantic-guided adversarial (SA) loss.

For the local adversarial loss, we first use refined patch group $P'$ as candidate fake patches of the output $I_{out}$. Then, we compare the discriminating result of image patches among $P'$, the worst patch is most likely to be ‘fake’ and can be chosen to update parameters of both discriminator and generator. Hence, the discriminator plausibly uses the semantic priors to find the target fake region $x_f\mathit{\sim} p_{fake}$ by itself. While the real patches $x_r \mathit{\sim} p_{real}$ are still randomly cropped from real images each time. The local adversarial loss function is defined as:
\vspace{-0.1cm}
\begin{equation}
\begin{split}
  \mathcal{L}_{local} = &\min\limits_G \max\limits_D\mathbb{E}_{x_r\sim p_{real}} MSE(D(x_r),0) \\
  &+ \mathbb{E}_{x_f\sim p_{fake}} MSE(D(x_f),1), 
  \label{eq:saloss1}
\end{split}
\end{equation}
\vspace{-0.1cm}
\begin{equation}
  x_f = P^t, D(P^t)=\min(D(P^0), \dotsc, D(P^{class})),
  \label{eq:saloss2}
  \vspace{-0.1cm}
\end{equation}

where $MSE(\cdot)$ denotes the mean squared error and $P^t$ denotes the target fake patch.

For the global adversarial loss, we adopt a simple design to achieve semantic-aware guidance when discriminating a fake sample. We concatenate $I_{out}$ and $I_{seg}'$, which is the output feature before $Softmax$, as a new $x_f$. The images $x_r$ with real distribution are randomly sampled. Finally, the global adversarial loss function is defined as:
\vspace{-0.1cm}
\begin{equation}
\begin{split}
  \mathcal{L}_{global} = &\min\limits_G \max\limits_D\mathbb{E}_{x_r\sim p_{real}} MSE(D(x_r),0) \\
  &+ \mathbb{E}_{x_f\sim p_{fake}} MSE(D(x_f, I_{seg}'),1), 
  \label{eq:saloss3}
  \vspace{-0.1cm}
\end{split}
\end{equation}

Therefore, the SA loss can be defined as:
\vspace{-0.1cm}
\begin{equation}
  \mathcal{L}_{SA}=\mathcal{L}_{global} + \mathcal{L}_{local},
  \label{eq:saloss4}
  \vspace{-0.1cm}
\end{equation}

We define the original loss function of Enhancement Net as $\mathcal{L}_{recon}$, which may be $l_1$ loss, $MSE$ loss, $SSIM$ loss, etc., or their combination according to original setting of each selected method. Thus, the overall loss function of our SKF can be formulated as follows:
\vspace{-0.1cm}
\begin{equation}
  \mathcal{L}_{all}=\mathcal{L}_{recon} + \lambda_{SCH}\mathcal{L}_{SCH} + \lambda_{SA}\mathcal{L}_{SA},
  \label{eq:saloss5}
  \vspace{-0.1cm}
\end{equation}

where $\lambda$s are weights to balance the loss terms. 

\vspace{-0.15cm}
\section{Experiments}
\vspace{-0.05cm}
\subsection{Experimental Settings}
\vspace{-0.05cm}

\textbf{Datasets.} We evaluate the proposed framework on several datasets from various scenes, including LOL~\cite{Chen2018Retinex}, LOL-v2~\cite{yang2021sparse}, MEF~\cite{ma2015mef}, LIME~\cite{guo2016lime-mm}, NPE~\cite{wang2013npe} and DICM~\cite{lee2013dicm}. The LOL dataset~\cite{Chen2018Retinex} is a real captured dataset including 485 low/normal light image pairs for training and 15 pairs for testing. The LOL-v2 dataset~\cite{yang2021sparse} is the real part of LOL-v2, which is larger and more diverse than LOL, including 689 low/normal light pairs for training and 100 pairs for testing. The MEF (17 images), LIME (10 images), NPE (85 images) and DICM (64 images) are real captured datasets including unpaired images.

\textbf{Metrics.} To evaluate the performance of different LLIE methods with and without our SKF, we use both full-reference and non-reference image quality evaluation metrics. For LOL/LOL-v2 datasets, peak signal-to-noise ratio (PSNR), structural similarity (SSIM)~\cite{wang2004ssim}, learned perceptual image patch similarity (LPIPS)~\cite{zhang2018lpips}, natural image quality evaluator (NIQE)~\cite{mittal2012niqe} are employed. For the MEF, LIME, NPE and DICM datasets without paired data, only NIQE is used, as there is no ground-truth.

\vspace{-0.01cm}
\textbf{Compared Methods.} To verify the effectiveness of our designs, we compare our method with a rich collection of SOTA methods for LLIE, including LIME~\cite{guo2016lime}, RetinexNet~\cite{Chen2018Retinex}, KinD~\cite{zhang2019kind}, DRBN~\cite{yang2020drbn}, KinD++~\cite{zhang2021kindplus}, Zero-DCE~\cite{guo2020zerodce}, ISSR~\cite{fan2020integrating}, EnlightGAN~\cite{jiang2021enlightengan}, MIRNet~\cite{zamir2022mirnetv2}, HWMNet~\cite{fan2022hwmnet}, SNR-LLIE-Net~\cite{xu2022snr}, LLFlow~\cite{wang2022llflow}. To demonstrate the superiority of our method faithfully, we reasonably select several methods as the baseline networks. Specifically, both the most representative methods including RetinexNet, KinD and KinD++, and three latest methods including HWMNet, SNR-LLIE-Net and LLFlow are selected. Thus, our methods are denoted as RetinexNet-SKF, KinD-SKF, DRBN-SKF, KinD++-SKF, HWMNet-SKF, SNR-LLIE-Net-SKF, LLFlow-S-SKF and LLFlow-L-SKF (small and large version of LLFlow respectively). 

\vspace{-0.01cm}
\textbf{Implementation Details.} We conduct our experiments on NVIDIA 3090 GPU and NVIDIA A100 GPU, which are based on the released code of the baseline networks with the same training settings. Specifically, only the last subnets of Retinex-SKF, KinD-SKF and KinD++-SKF are trained with SCH loss and SA loss, while the other subnets are trained with the original loss functions. Furthermore, we do not apply SA loss to LLFlow because there is no enhanced output in training stage. Additionally, SE modules are reasonably located in decoders of all the baseline networks.

\vspace{-0.17cm}
\subsection{Quantitative Evaluation}
\begin{figure*}[t]
  \centering
   \includegraphics[width=\linewidth]{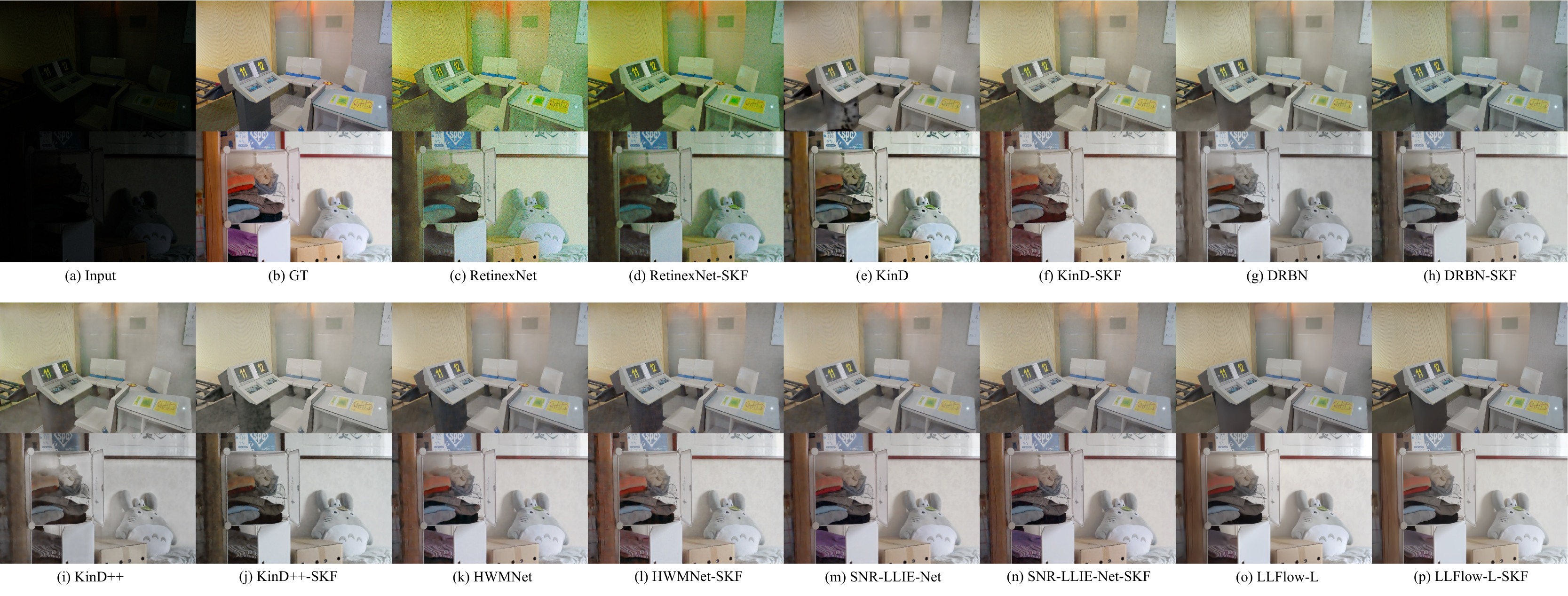}
   \setlength{\abovecaptionskip}{-0.4cm}
   \setlength{\belowcaptionskip}{-0.3cm}
   \caption{Visual comparison of baseline methods with and without SKF on LOL dataset. Our SKF enables baseline methods produce images with less noise, more color information and realistic details.}
   \label{fig:LOL_vis}
\end{figure*}

\begin{figure*}[t]
  \centering
   \includegraphics[width=\linewidth]{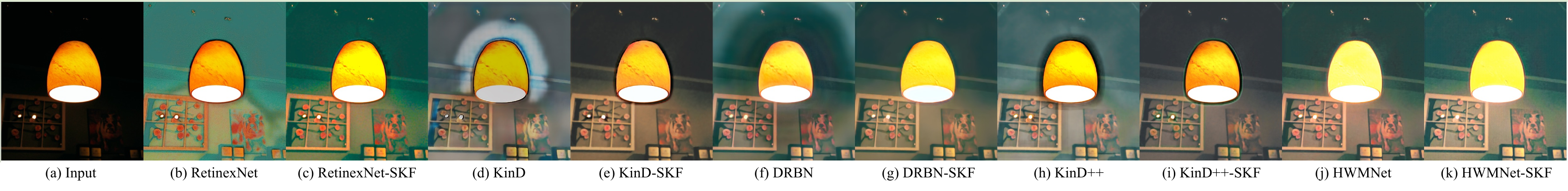}
   \setlength{\abovecaptionskip}{-0.4cm}
   \setlength{\belowcaptionskip}{-0.6cm}
   \caption{Visual comparison of baseline methods with and without SKF on LIME dataset.}
   \label{fig:LIME_vis}
\end{figure*}

\vspace{-0.1cm}
\textbf{Quantitative results on LOL and LOL-v2 datasets.} The evaluation results are shown in~\cref{tab:lollolv2}. We can observe that our SKF achieves consistent and significant performance gain over each baseline method. Specifically, our SKF provides an average improvement of 1.750 dB/1.611 dB on LOL/LOL-v2 datasets respectively and this is achieved by introducing the capability of suppressing noise and artifacts and preserving color consistency. Notably, our LLFlow-L-SKF earns PSNR values of 26.798 dB/28.451 dB on LOL/LOL-v2 datasets, establishing a new SOTA. Furthermore, SSIM values achieve similar performance as well. Our SKF yields better SSIM values by an average of 0.041/0.037 on LOL/LOL-v2 datasets, which illustrates that our SKF helps baseline methods restore the luminance and contrast and preserve the structural information with details. Besides, the substantial gain of LPIPS and NIQE provided by our SKF reasonably indicates that human intuition is more closely matched by introducing semantic priors from our designs. 

% Table generated by Excel2LaTeX from sheet 'MEF'
\begin{table}[t]
  \centering
  \setlength{\abovecaptionskip}{0.15cm}
  \caption{Quantitative comparison on the LOL~\cite{Chen2018Retinex}, LOL-v2~\cite{yang2021sparse}, MEF~\cite{ma2015mef}, LIME~\cite{guo2016lime-mm}, NPE~\cite{wang2013npe} and DICM~\cite{lee2013dicm} datasets in terms of NIQE, where smaller values lead to better quality.}
  \scalebox{.66}{
    \begin{tabular}{c|c|c|c|c|c|c}
    \toprule
    \textbf{Method} & \textbf{LOL} & \textbf{LOL-v2} & \textbf{MEF} & \textbf{LIME} & \textbf{NPE} & \textbf{DICM} \\
    \midrule
    Input & 6.7488 & 6.7911 & 4.2650 & 4.4380 & 4.3124 & 4.2550 \\
    \midrule
    \midrule
    RetinexNet~\cite{Chen2018Retinex} \scriptsize BMVC'18 & 6.8731 & 5.8488 & 4.1490 & 4.4200 & 4.5008 & 4.5912 \\
    \textbf{RetinexNet-SKF(Ours)} & 4.2118 & 4.2331 & \textbf{3.6321} & 4.0779 & 4.0152 & 3.6945 \\
    \midrule
    \midrule
    KinD~\cite{zhang2019kind} \scriptsize MM'19 & 5.1891 & 6.8490 & 4.1344 & 4.6418 & 4.6896 & 3.9371 \\
    \textbf{KinD-SKF(Ours)} & 5.0306 & 4.7783 & 3.9460 & 4.3607 & 3.8721 & 3.7909 \\
    \midrule
    \midrule
    DRBN~\cite{yang2020drbn} \scriptsize CVPR'20  & 4.7930 & 4.9612 & 4.0956 & 4.4019 & 3.9205 & 4.0433 \\
    \textbf{DRBN-SKF(Ours)} & 4.4636 & 4.4599 & 4.0894 & 4.3392 & 4.0192 & 3.8541 \\
    \midrule
    \midrule
    KinD++~\cite{zhang2021kindplus} \scriptsize IJCV'20 & 4.7602 & 5.0856 & 3.7498 & 4.3756 & 3.9848 & 3.7076 \\
    \textbf{KinD++-SKF(Ours)} & \textbf{4.1415} & \textbf{4.1785} & 3.7645 & \textbf{3.9892} & \textbf{3.8201} & \textbf{3.5382} \\
    \midrule
    \midrule
    HWMNet~\cite{fan2022hwmnet} \scriptsize ICIP'22 & 5.1407 & 5.9702 & 4.2175 & 4.3549 & 4.0683 & 3.9196 \\
    \textbf{HWMNet-SKF(Ours)} & 4.3460 & 4.6826 & 4.0312 & 4.3699 & 3.9942 & 4.0760 \\
    \bottomrule
    \end{tabular}}%
  \label{tab:niqe}%
  \vspace{-0.6cm}
\end{table}%

\textbf{Quantitative results on MEF, LIME, NPE and DICM datasets.} The evaluation results on the MEF, LIME, NPE and DICM datasets are described in~\cref{tab:niqe}. In general, each method with SKF obtain better NIQE results than baseline on all six datasets except three worse cases of DRBN-SKF and HWMNet-SKF.\, The RetinexNet-SKF performs the best with NIQE of 3.632 on MEF dataset, while the KinD++-SKF achieves the best performance on other five datasets. Overall, it is notable that our SKF yields an average gain of 0.519 on NIQE across all the methods and datasets. The better NIQE shows that the methods with our SKF can produce images with more natural textures and become more effective for restoring low-light images. 

\vspace{-0.2cm}
\subsection{Qualitative Evaluation}
\vspace{-0.2cm}
The qualitative evaluations on LOL and LIME datasets are shown in~\cref{fig:LOL_vis,fig:LIME_vis} respectively. As indicated by~\cref{fig:LOL_vis}, our SKF can improve the enhancement capability of baseline methods and generate images with more pleasing perceptual quality. Specifically, the results of RetinexNet are unreal due to the obvious color gap and serious noise, which can be mitigated by our SKF. Compared to results of KinD and KinD++, KinD-SKF and KinD++-SKF resolve the issue of inconsistent lighting and strange white artifacts. For other results, more consistent color and natural details recovery for desk, wall and clothes are achieved by our SKF.

\begin{figure*}[ht]
  \centering
  \includegraphics[width=\linewidth]{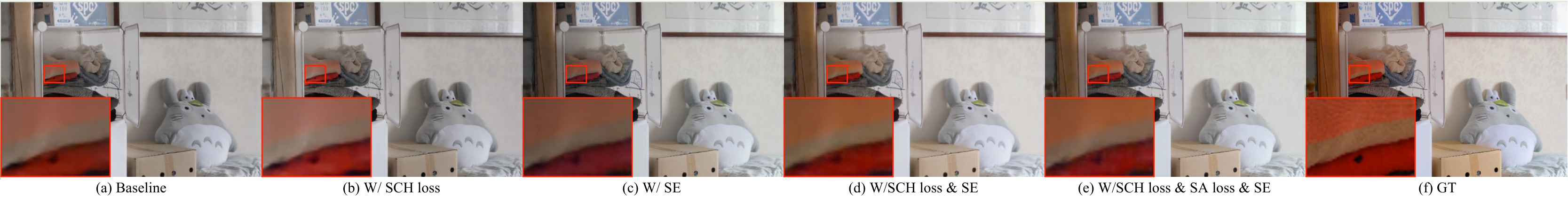}
  \setlength{\abovecaptionskip}{-0.3cm}
  \setlength{\belowcaptionskip}{-0.2cm}
  \caption{Visual comparison of DRBN-SKF for investigating the contribution of key techniques of our SKF.}
  \label{fig:abl_vis}
\end{figure*}

% Table generated by Excel2LaTeX from sheet 'Ablation_all'
\begin{table*}[ht]
  \centering
  \setlength{\abovecaptionskip}{0.15cm}
  \caption{Ablation study of KinD++-SKF, DRBN-SKF and HWMNet-SKF for investigating the contribution of key techniques of our SKF.}
  \scalebox{.68}{
    \begin{tabular}{ccc||c|c|c|c||c|c|c|c||c|c|c|c}
    \toprule
    \multirow{2}[2]{*}{\textbf{SCH loss}} \quad & \multirow{2}[2]{*}{\textbf{SA loss}} \quad & \multirow{2}[2]{*}{\textbf{SE module}} & \multicolumn{4}{c||}{\textbf{KinD++-SKF}} & \multicolumn{4}{c||}{\textbf{DRBN-SKF}} & \multicolumn{4}{c}{\textbf{HWMNet-SKF}} \\
\cmidrule{4-15}          &       &       & \textbf{PSNR} $\uparrow $ & \textbf{SSIM} $\uparrow $ & \textbf{LPIPS} $\downarrow $ & \textbf{NIQE} $\downarrow $ & \textbf{PSNR} $\uparrow $ & \textbf{SSIM} $\uparrow $ & \textbf{LPIPS} $\downarrow $ & \textbf{NIQE} $\downarrow $ & \textbf{PSNR} $\uparrow $ & \textbf{SSIM} $\uparrow $ & \textbf{LPIPS} $\downarrow $ & \textbf{NIQE} $\downarrow $ \\
    \midrule
    & & &  18.970  & 0.804  & 0.175  & 4.760  & 19.860  & 0.834  & 0.155  & 4.793  & 24.240  & 0.852  & 0.114  & 5.141  \\
    \checkmark & & & 19.170  & 0.806  & \textbf{0.170}  & 4.759  & 20.040  & 0.835  & 0.154  & 4.793  & 24.590  & 0.859  & 0.112  & 5.023  \\
    & \checkmark & & 19.385  & 0.800  & 0.189  & 4.392  & 20.070  & 0.834  & 0.149  & 4.701  & 24.305  & 0.853  & 0.111  & 4.712  \\
    & & \checkmark & 19.781  & 0.808  & 0.181  & 4.712  & 21.334  & 0.837  & 0.143  & 4.678  & 24.477  & 0.859  & 0.111  & 4.988  \\
    \checkmark & & \checkmark & \textbf{20.620}  & \textbf{0.815}  & 0.176  & 4.536  & 22.550  & 0.836  & 0.150  & 4.581  & \textbf{25.123}  & 0.860  & 0.111  & 4.711  \\
    \checkmark & \checkmark & \checkmark & 20.363 & 0.805 & 0.201 & \textbf{4.142} & \textbf{22.837} & \textbf{0.841} & \textbf{0.138} & \textbf{4.464} & 25.086 & \textbf{0.860} & \textbf{0.108} & \textbf{4.346} \\
    \bottomrule
    \end{tabular}}%
    \label{tab:ablation_all}%
    \vspace{-0.5cm}
\end{table*}%

% Table generated by Excel2LaTeX from sheet 'Ablation_loss'
\begin{table}[ht]
  \centering
  \setlength{\abovecaptionskip}{0.15cm}
  \caption{Ablation study of HWMNet-SKF for investigating the effect of semantic priors in the loss function.}
  \scalebox{.63}{
    \begin{tabular}{cc||ccc||c|c|c|c}
    \toprule
    \multicolumn{2}{c||}{$\mathcal{L}_{SCH}$} & \multicolumn{3}{c||}{$\mathcal{L}_{SA}$} & \multirow{2}[1]{*}{\textbf{PSNR} $\uparrow $} & \multirow{2}[1]{*}{\textbf{SSIM} $\uparrow $}  & \multirow{2}[1]{*}{\textbf{LPIPS} $\downarrow $}  & \multirow{2}[1]{*}{\textbf{NIQE} $\downarrow $} \\
    w/o S & w/ S & w/o SA & w/o S  &w/ S & & &\\
    \midrule
    &       &       &       &       &24.240  & 0.852  & 0.114 & 5.141 \\
    \checkmark &       &\checkmark       &       &      &24.477  & 0.859  & 0.111 & 4.988  \\
    \checkmark &       &       & \checkmark      &      &24.568  & 0.857  & 0.113 & 4.613 \\
    \checkmark &       &       &       &\checkmark      &24.668  & 0.859  & 0.111 & 4.567 \\
    \midrule
    &\checkmark       & \checkmark      &  &         &\textbf{25.123}  & \textbf{0.860}  & \textbf{0.108} & 4.711 \\
    &\checkmark       &       &\checkmark       &    &25.040  & 0.859  & 0.111 & 4.546 \\
    &\checkmark       &       &       & \checkmark   &25.086  & \textbf{0.860}  & \textbf{0.108} & \textbf{4.311} \\
    \bottomrule
    \end{tabular}}%
  \label{tab:ablation_loss}%
  \vspace{-0.3cm}
\end{table}%

% Table generated by Excel2LaTeX from sheet 'Ablation_Param'
\begin{table}[ht]
  \centering
    \setlength{\abovecaptionskip}{0.15cm}
  \caption{Ablation study for investigating whether the performance improvement comes from semantic priors or more parameters.}
  \scalebox{.67}{
    \begin{tabular}{c|c||c|c|c|c}
    \toprule
    \multicolumn{2}{c||}{Method}      & \textbf{PSNR}  $\uparrow $    & \textbf{SSIM}  $\uparrow $    & \textbf{LPIPS} $\downarrow $     & \textbf{Param(M)} \\
    \midrule
    \multirow{3}[1]{*}{HWMNet} & Baseline   & 24.240  & 0.852  & 0.114  & 66.56  \\
     & Large     & 24.445  & 0.853  & 0.115  & 69.99  \\
     & w/ SKF    & \textbf{25.086}  & \textbf{0.860}  & \textbf{0.108}  & 69.98  \\
    \midrule
    \multirow{3}[1]{*}{LLFlow-S} & Baseline & 24.060  & 0.860  & 0.136  & 4.97  \\
     & Large     & 24.167  & 0.858  & 0.137  & 5.38  \\
     & w/ SKF      & \textbf{25.942}  & \textbf{0.865}  & \textbf{0.125}  & 5.26  \\
    \midrule
    \multirow{3}[1]{*}{LLFlow-L} & Baseline & 24.999  & 0.870  & 0.117  & 37.68  \\
     & Large     & 25.292  & 0.873  & 0.113  & 40.55  \\
     & w/ SKF      & \textbf{26.798}  & \textbf{0.879}  & \textbf{0.105}  & 39.91  \\
    \bottomrule
    \end{tabular}}%
  \label{tab:ablation_msa}%
  \vspace{-0.6cm}
\end{table}%

We further exhibit the visual enhancement results on the LIME dataset in~\cref{fig:LIME_vis}. It can be observed that the methods with our SKF suppress the unnatural halo around the lamp and restore naturalistic color and details. Hence, methods with our SKF yield more visually pleasing results as compared to baselines, supporting our method's excellent performance in quantitative evaluation. More visualization results are provided in the \textcolor[rgb]{0.75,0.16,0.26}{supplementary material}.

\vspace{-0.2cm}
\subsection{Ablation Study}
\vspace{-0.2cm}
We conduct ablation studies on LOL dataset to prove the effectiveness of our SKF from various aspects. 
% \lichongyi{It would be great if we could show some visual results of ablated models.}

\textbf{SCH loss, SA loss and SE module.} As shown in~\cref{tab:ablation_all}, we conduct experiments of KinD++-SKF, DRBN-SKF and HWMNet-SKF. The addition of SCH loss and SE module improve the PSNR by an average of 0.243 dB and 0.841 dB over the baseline respectively. Simultaneously applying SCH loss and SE module further improves the baseline method by yielding an average gain of 1.741 dB over the baseline. This verifies that more beneficial semantic-aware priors are integrated into enhancement process. Despite that adding SA loss causes minor drops in some full-reference metrics, average gain of 0.292 with NIQE are obtained across all the cases. Therefore, the baseline methods are refined through semantic-aware knowledge by each component, and the total framework leads to a significant performance boost. Additionally, results in~\cref{fig:abl_vis} demonstrate that model with SCH loss and SE module can preserve color consistency and details and the SA loss reduces fake regions by producing more natural textures. 

\textbf{Semantic-guided losses.}~\cref{tab:ablation_loss} lists the results of different settings of losses. The w/o S and w/ S denote calculating the global histogram and our semantic-guided histogram respectively. For SA loss, the w/o SA, w/o S and w/ S denote without SA loss, classic global and local adversarial loss like EnlightGAN~\cite{jiang2021enlightengan} and our SA loss. First, the HWMNet-SKF with SCH loss presents better performance, achieving average margin of 0.512 dB improvement on PSNR, indicating the significant capability of SCH loss in preserving color consistency. Furthermore, the average gain of 0.271 on NIQE by adding classic adversarial loss can be attributed to the capability of discriminator to improve visual quality. Finally, our SA loss provides favorable gain of 0.411 on NIQE over the baseline and faithfully demonstrates that the semantic priors help find out fake regions and thus produce more natural images.

\textbf{Superiority of semantic priors.} We choose HWMNet-SKF, LLFlow-S-SKF and LLFlow-L-SKF to investigate whether the improvement of performance benefits from semantic priors provided by our SKF or the more parameters of our SE modules. As shown in~\cref{tab:ablation_msa}, Baseline, Large and w/ SKF denote the original model, original model with more layers or channels and original model with our SKF. Our methods achieve significant improvement by average margin of 1.272 dB on PSNR comparing to large versions with similar number of parameters. Hence, we prove the superiority of the semantic priors instead of extra parameters. 

\vspace{-0.1cm}
\section{Conclusion}
\vspace{-0.1cm}
This work has proposed a novel framework for semantic-aware image enhancement, named SKF. The SKF incorporates semantic priors into Enhancement Net to preserve color consistency and visual details by SE module, SCH loss and SA loss. SE module allows image features to perceive rich and spatial information by semantic feature representations. SCH loss offers effective semantic-aware regional constrain for preserving color consistency. SA loss combines global and local adversarial loss and semantic priors to seek target fake region and generates natural results. Extensive experiments show that our SKF achieves superior performance in the case of all six baseline methods, and the LLFlow-L-SKF outperforms all the competitors. However, the improvement is limited when dealing with unknown category, inducing more possibility when improving the capability of identifying unknown instance by SKB. Furthermore, we will also explore the potential of our SKF in other low-level vision tasks.

\vspace{0.1cm}
\noindent \textbf{Acknowledgements:} This work was supported in part by the National Natural Science Foundation of China under grant 62102069, U20B2063 and 62220106008, the Sichuan Science and Technology Program under grant 2022YFG0032, and the China Academy of Space Technology (CAST) Innovation Program. We are also sponsored by
CAAI-Huawei MindSpore Open Fund.

%%%%%%%%% REFERENCES
{\small
\bibliographystyle{ieee_fullname}
\bibliography{mybib}
}

\end{document}